\documentclass{article}
\usepackage{spconf,amsmath,graphicx}
\usepackage{multirow,enumerate,makecell,subfigure}
\usepackage{amssymb}

\title{BiSVP: Building Footprint Extraction via Bidirectional Serialized Vertex Prediction}

\name{Mingming Zhang$^{1}$, Ye Du$^{1}$, Zhenghui Hu$^{2}$, Qingjie Liu$^{\star1}$, Yunhong Wang$^{1}$}
\address{$^{1}$ State Key Laboratory of Virtual Reality Technology and Systems, Beihang University, Beijing, China \\
$^{2}$ Hangzhou Innovation Institute, Beihang University}

\begin{document}
%\ninept
%
\maketitle
\begin{abstract}
\small
Extracting building footprints from remote sensing images has been attracting extensive attention recently. Dominant approaches address this challenging problem by generating vectorized building masks with cumbersome refinement stages, which limits the application of such methods. In this paper, we introduce a new refinement-free and end-to-end building footprint extraction method, which is conceptually intuitive, simple, and effective. Our method, termed as BiSVP, represents a building instance with ordered vertices and formulates the building footprint extraction as predicting the serialized vertices directly in a bidirectional fashion. Moreover, we propose a cross-scale feature fusion (CSFF) module to facilitate high resolution and rich semantic feature learning, which is essential for the dense building vertex prediction task. Without bells and whistles, our BiSVP outperforms state-of-the-art methods by considerable margins on three building instance segmentation benchmarks, clearly demonstrating its superiority. The code and datasets will be made public available.
\end{abstract}
\section{Introduction}
\label{sec:intro}

Extracting building footprints from remote sensing images has been receiving increasing attention due to its great potential value in many applications, such as urban change detection, city modeling, and cartography, which require precise geometric contours. Most prevalent methods address this task by vectorizing building segmentation masks, heavily relying on the performance of segmentation methods. Another line of works directly predict the order-agnostic vertex set, however, they require additional information to determine the vertex's order. In a nutshell, mainstream approaches tend to have complex model structures and tedious inference processes.

To tackle the above issues, we propose a Bidirectional Serialized Vertex Prediction (BiSVP) framework to predict the \textit{ordered sequence} of building vertices. The method is refinement-free and can be trained end-to-endly. Considering that a polygon can be represented with sequential vertices in clockwise or counterclockwise direction, our BiSVP represents the building contour with ordered vertices and predicts building vertices sequentially in a bidirectional fashion. The predictions from two directions are then combined to produce the final results. In this way, we manage to leverage the \textit{bidirectional} information of building polygons to generate accurate building footprints. Besides, an attention mechanism is integrated into our BiSVP to enhance the ability of predicting long vertex sequences of complex buildings. Furthermore, we propose a cross-scale feature fusion (CSFF) module to obtain building features with high resolution and rich semantic information.

Our BiSVP can be seamlessly incorporated into existing object detectors (e.g., Faster RCNN \cite{ren2015fasterrcnn}). Experimental results show that our method significantly outperforms state-of-the-art approaches on three building instance segmentation benchmarks. In short, the contributions are summarized as follows: First, we propose Bidirectional Serialized Vertex Prediction, a simple yet effective end-to-end framework to extract building footprints. Second, our method predicts the serialized vertices directly in a bidirectional fashion and proposes the cross-scale feature fusion module to enhance the building feature learning. Third, our method outperforms state-of-the-art approaches by considerable margins on three building instance segmentation benchmarks.

\vspace{-10pt}
\section{Related Work}
\label{sec:format}

In this section, we review literature closely related to our research.
\vspace{-10pt}
\subsection{Building instance segmentation}
Early works address building footprint extraction as a pixel-wise classification problem. To improve the segmentation quality, multi-source information, such as digital surface data and LIDAR data, are incorporated to obtain rich features \cite{awrangjeb2010automatic,li2018semantic}. In the era of deep learning, numerous instance-level building segmentation methodologies have emerged \cite{li2018hough,zhao2018building,zorzi2019regularization,xu2021gated}, which benefit a lot from instance segmentation networks. However, these methods provide building masks in raster format, which can not meet application needs.
\begin{figure*}[htb]
\centering
\includegraphics[width=0.8\textwidth,height=0.5\textwidth]{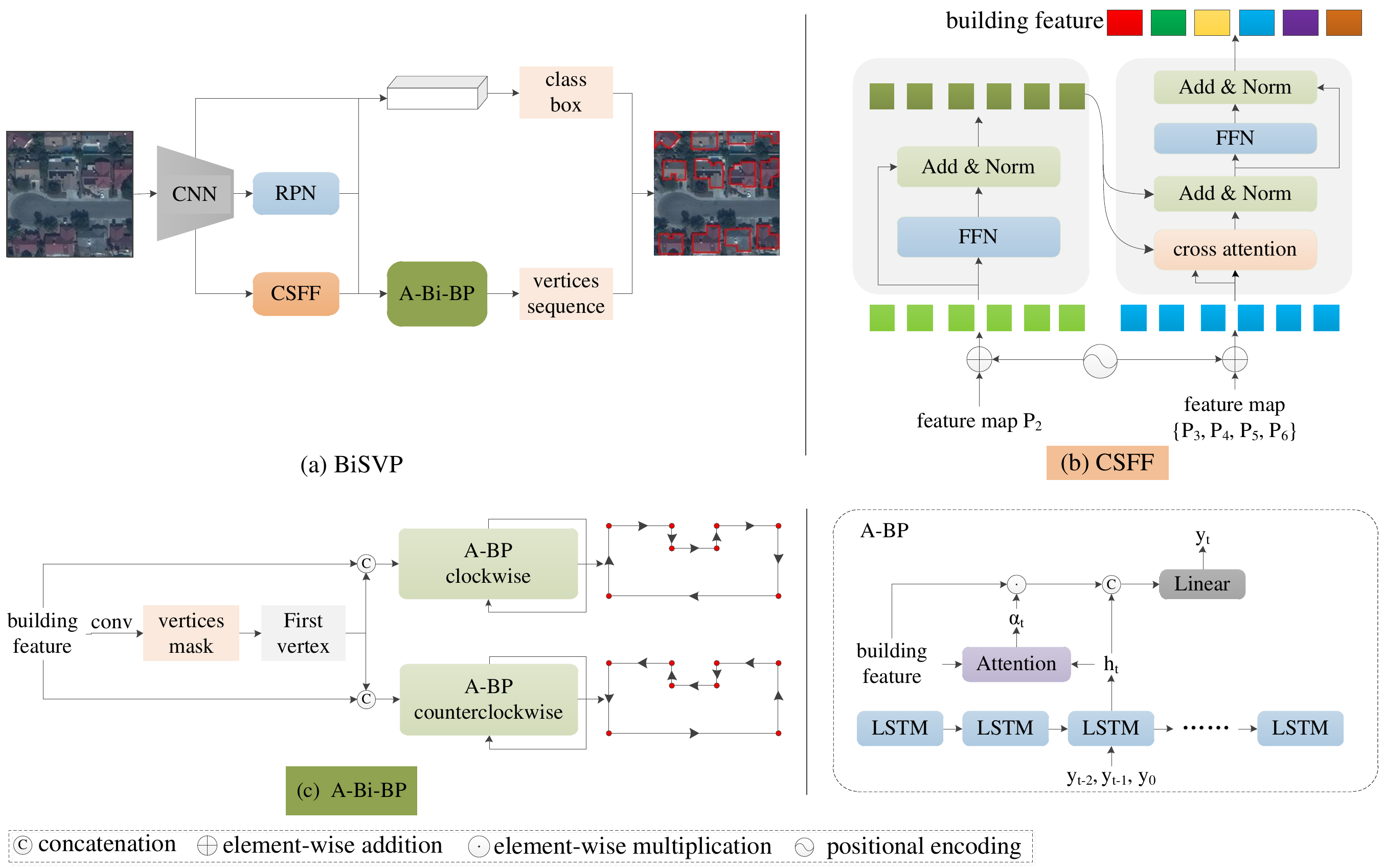}
\caption{Overview of our proposed BiSVP. (a) BiSVP is a refinement-free and end-to-end framework. (b) Cross-scale feature fusion (CSFF) module is introduced to facilitate high resolution and rich semantic feature learning. (c) Attentional bidirectional building polygon (A-Bi-BP) prediction module is proposed to predict the serialized vertices in a bidirectional fashion directly.}
\label{Fig2}
\end{figure*}
\vspace{-10pt}
\subsection{Polygonal building segmentation}
Polygonal building segmentation approaches extract building footprints in a vector format. Most current methods \cite{chen2020polygoncnn,liang2020polytransform,li2020approximating,zorzi2021machine,chen2021quantization} vectorize building segmentation masks by post-processing. Girard et al. \cite{girard2021framefield} predicts raster segmentation and frame fields to generate building polygons. These methods heavily rely on building segmentation masks and consequently generate irregular building polygons.

Another mainstream algorithms focus on directly predicting building vertices from the feature maps. Li et al. \cite{li2020instance} and Zhu et al. \cite{zhu2021apga} determine the building vertex's order in a geometric way. PolyWorld \cite{zorzi2021polyworld} assigns the vertex connections for building vertices by solving a differentiable optimal transport problem. These methods decompose the polygonal building segmentation problem into multiple tasks and require complex polygon constraints, normally bringing intensive computation burden and leading to poor generalization. To determine the vertex's order, PolyRNN \cite{castrejon2017polygonrnn} and PolyRNN++ \cite{acuna2018polygonrnn++} apply a CNN-RNN architecture to directly extract vertex sequences. Following this simple idea, many attempts \cite{li2019polymapper,zhao2021building} have been made and achieved promising results. However, these methods are sensitive to inevitable occlusions and shadows, which may be challenging to extract complex building vertices.

\section{Method}
\label{sec:pagestyle}

The overall architecture of BiSVP is shown in Figure \ref{Fig2}, including a feature extraction network, a cross-scale feature fusion (CSFF) module, and an attentional bidirectional building polygon (A-Bi-BP) prediction module.
\vspace{-10pt}
\subsection{Feature Extraction Network}
Our method can be seamlessly embedded into any polygonal models. To evaluate the effectiveness of it, we build a model on top of Mask RCNN because of its popularity and excellence performance. To be specific, BiSVP adopts a deep CNN backbone to extract multi-scale features $\textit{F}$. To improve multi-scale building segmentation, we apply a feature pyramid network (FPN) \cite{lin2017fpn} to fuse feature maps of different resolutions \{$\textit{P}_2, \textit{P}_3, \textit{P}_4, \textit{P}_5, \textit{P}_6$\}. Then, a region proposal network (RPN) \cite{ren2015fasterrcnn} proposes candidate building bounding boxes from the multi-scale features.
\vspace{-10pt}
\subsection{Cross-scale Feature Fusion}
As depicted in Figure \ref{Fig2} (b), CSFF adopts the transformer-based architecture to obtain the high resolution and rich semantic building representations, which are crucial for the subsequent vertex sequence prediction. Especially, CSFF primarily employs the cross attention module to aggregate feature maps of different resolutions in a coarse to fine manner, which can automatically focus on building boundaries.

Given an image $\textit{I}\in R^{3 \times H \times W}$, CSFF firstly obtains the building queries $\textit{B}_{q}$ from the feature map $\textit{P}_2$ and the positional encoding. Then, CSFF takes in feature maps ${\textit{P}_3, \textit{P}_4, \textit{P}_5, \textit{P}_6}$, which embeds positional encoding with the corresponding feature map to localize building instances and boundaries in the following cross attention. The cross attention module aggregates information between building queries $\textit{B}_{q}$ and feature maps $\textit{P}_i$ ($i \in [3,4,5,6]$). Besides, the residual connection and layer normalization are also applied after the attention and FFN modules. The output of the CSFF module is defined as:
\begin{equation} \label{eqn1}
  B = \text{FFN}(\text{softmax}(\frac{B_{q}P_i^{k}}{\sqrt{d}})\cdot{P_i^{v}})
\end{equation}
where $B_{q}$ is the building queries from the feature map $\textit{P}_2$, and $\text{FFN}$ is the fully connected feed-forward network. $P_i^{k}$ and $P_i^{v}$ represent key and value from feature map $P_i$ ($i \in [3,4,5,6]$) from FPN.
\vspace{-10pt}
\subsection{Attentional bidirectional building polygon}
As illustrated in Figure \ref{Fig2} (c), A-Bi-BP module firstly takes in the building feature $\textit{B}$ to get the first vertex. Then, it outputs building vertices in two directions (i.e., clockwise and counterclockwise). In the following, we take step $\textit{t}$ as an example to introduce the attentional building polygon (A-BP) prediction, which is a branch of A-Bi-BP module.

Firstly, A-BP module outputs the hidden state $h_t$ by taking in the building feature $\textit{B}$, the previous predicted vertices $y_{t-2}$ and $y_{t-1}$, and the predicted first vertex $y_0$. Then, a gaussian constrained attention \cite{qiao2021gaussian} module integrating the hidden state $h_t$ and the building feature $\textit{B}$ is applied to calculate the attention weight ${\alpha}_t$. Subsequently, the coefficient is calculated by the element-wise product of the attention weight ${\alpha}_t$ and the building feature $\textit{B}$. Finally, the next vertex $y_t$ or the end signal (EOS) is captured from the concatenation of the coefficient and the hidden state $h_t$. The step t of the A-BP is defined as:
\begin{equation} \label{eqn2}
  \begin{split}
  (h_t, c_t)&=\text{LSTM}(B, y_{t-2}, y_{t-1}, y_0),    \\
  y_t&=\text{softmax}(W[h_t;{{\alpha}_t}\odot{B}]+b)
  \end{split}
\end{equation}
where ${\alpha}_t$ is $\text{attn}_{gc}(B, h_t)$, ${\alpha}_t\odot{B}$ is the element-wise product of the attention weights and $B$, and $[;]$ is the concatenation operation. $W$ and $b$ are the trainable parameters and $y_t$ is obtained by $\text{softmax}$ operation.
\vspace{-10pt}
\subsection{Training objective}
BiSVP loss includes binary classification loss $L_{cls}$, bounding box regression loss $L_{reg}$, and building vertex sequence prediction loss $L_{ver}$. This paper adopts the binary cross entropy loss and L1 loss to calculate $L_{cls}$ and $L_{reg}$, respectively. As for $L_{ver}$, we calculate the average loss of the cross entropy loss between predicted polygons of two directions with the corresponding ground truth respectively. The total loss is defined as follows:
\begin{equation} \label{eqn3}
  \begin{split}
  L_{ver}&=(\text{L}_{ce}(pred_p, gt) + \text{L}_{ce}(pred_p^{c}, gt^{c})) / 2.0, \\
  L&=L_{cls} + L_{reg} + L_{ver}.
  \end{split}
\end{equation}
where $pred_p$ and $gt$ ($pred_p^{c}$ and $gt^{c}$) represent the predicted building polygon and ground truth in clockwise (counterclockwise).

\section{Experiments}
\label{sec:typestyle}
\vspace{-10pt}
\subsection{Datasets}
The proposed method is evaluated on three building datasets: (1) SpaceNet (LasVegas) \cite{van2018spacenet} consists of over 3,800 images of size 650$\times$650 pixels. (2) 5M-Building \cite{lu20195m} contains 109 images with a resolution ranging from 2000×2000 to 5000×5000. We crop them into $512\times512$ sub-images with an overlap of 64 pixels and then split it by 7:3 for training and testing. (3) CNData is a very challenging dataset, including 4200 images with a size of 512$\times$512. The images are captured over different provinces of China and consist of residential, rural and industrial areas, where buildings vary greatly in size, structure and appearance. Especially in rural and urban villages, buildings are small and dense. The dataset has 101430 building instances with polygonal annotations, which is split by 8:1:1 for training, validation and testing.
\vspace{-10pt}
\subsection{Implementation Details}
Our proposed model is trained in an end-to-end manner with SGD \cite{robbins1951sgd} optimizer. The backbone is a ResNet50 pre-trained on ImageNet, and we fine-tune it with a smaller initial learning rate 1e-5; for the other part of the model, the initial learning rate is set to 1e-4. The weight decay is set to 1e-4. The model is trained for 24 epochs, and we decrease the learning rate by 10 at the 16-th and 22-th epoch, respectively. We use MS COCO metrics \cite{lin2014mscoco} to evaluate the segmentation results. Furthermore, $F1_{75}$ calculated from $AP_{75}$ and $AR_{75}$ is also employed to comprehensively 
evaluate different methods, since high accuracy delineation is vital for practical applications.
\vspace{-10pt}
\subsection{Comparison with State-of-the-arts}
Since building footprint extraction is also a instance segmentation task, we compare it with the baseline model Mask R-CNN \cite{he2017maskrcnn} and PANet \cite{liu2018panet}. Furthermore, we also compare the proposed BiSVP with PolyMapper \cite{li2019polymapper} and the SOTA method Framefield \cite{girard2021framefield} for polygonal building segmentation.

\noindent \textbf{Quantitative Evaluation}.
Table \ref{tab1} reports the building segmentation results on three building datasets. As illustrated in Table \ref{tab1}, $F1_{75}$ of our method on the three datasets are significantly better than the baseline method by 13.34\%, 2.93\%, and 5.15\%; $AP_{75}$ are improved by 15.7\%, 2.1\%, and 4\%, which indicates that our method can extract building footprint precisely. Besides, the recall metrics are all enhanced by our approach on three building test datasets, especially, +9.6\% in terms of $AR_{75}$ on SpaceNet. Moreover, our model outperforms the polygonal building segmentation methods by large margins. 

\begin{figure}[htbp]
	\subfigure{
        \rotatebox{90}{\scriptsize{~~~~~~~~~~GT}}
		\begin{minipage}[t]{0.23\linewidth}
			\centering
			\includegraphics[width=1\linewidth]{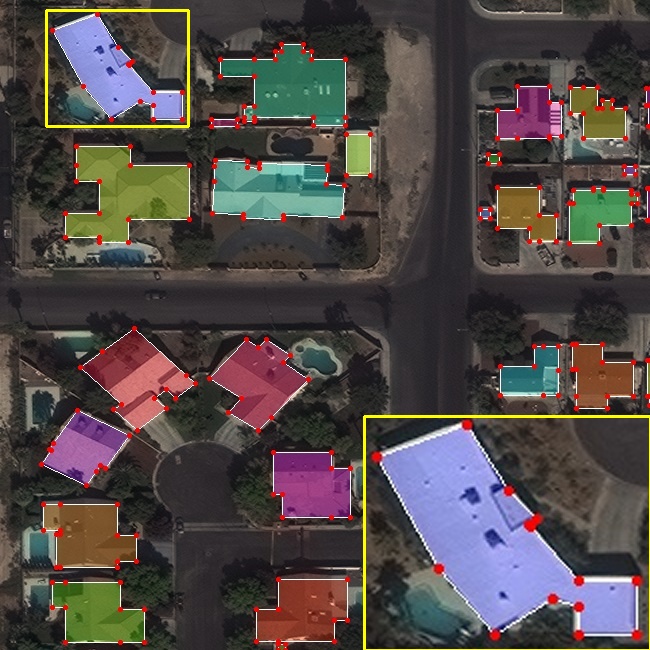}
		\end{minipage}
		\begin{minipage}[t]{0.23\linewidth}
			\centering
			\includegraphics[width=1\linewidth]{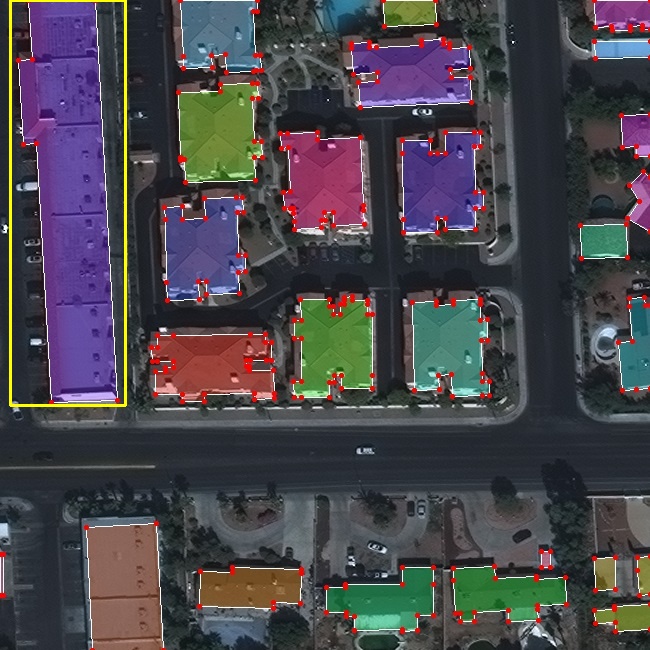}
		\end{minipage}
		\begin{minipage}[t]{0.23\linewidth}
			\centering
			\includegraphics[width=1\linewidth]{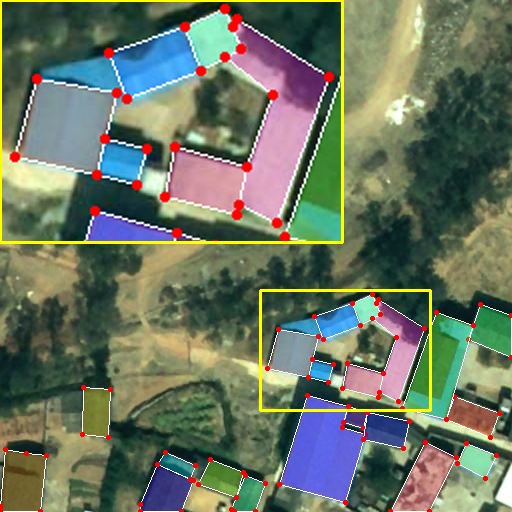}
		\end{minipage}
		\begin{minipage}[t]{0.23\linewidth}
			\centering
			\includegraphics[width=1\linewidth]{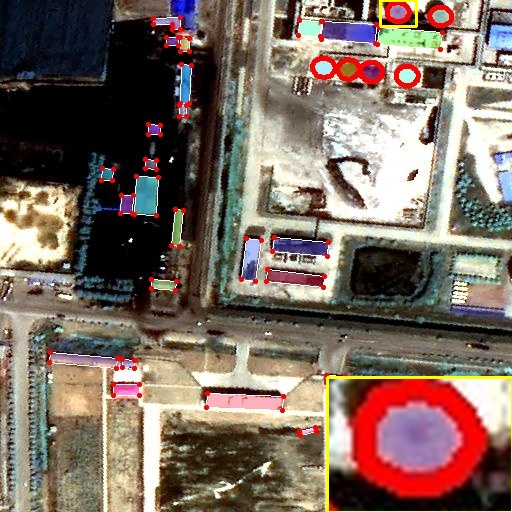}
		\end{minipage}
	}

	\vspace{-2mm}
	\setcounter{subfigure}{0}

    \subfigure{
		\rotatebox{90}{\scriptsize{~~~~~~~~BiSVP}}
		\begin{minipage}[t]{0.23\linewidth}
			\centering
			\includegraphics[width=1\linewidth]{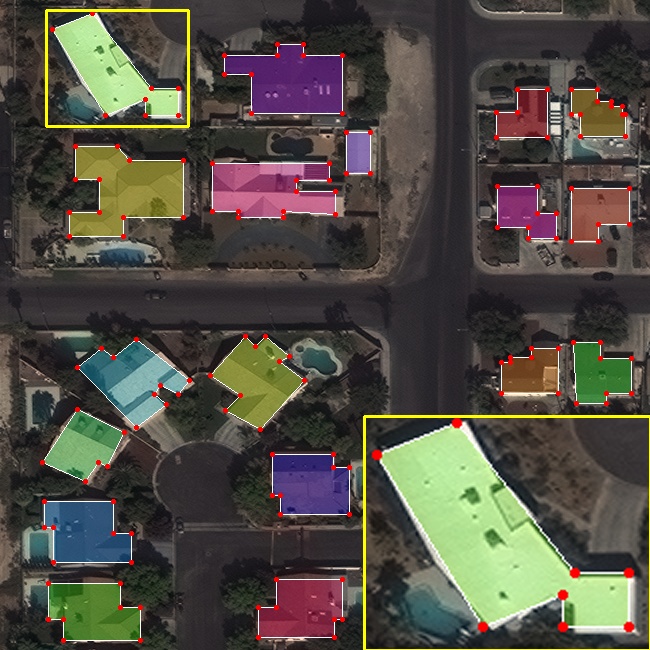}
		\end{minipage}
		\begin{minipage}[t]{0.23\linewidth}
			\centering
			\includegraphics[width=1\linewidth]{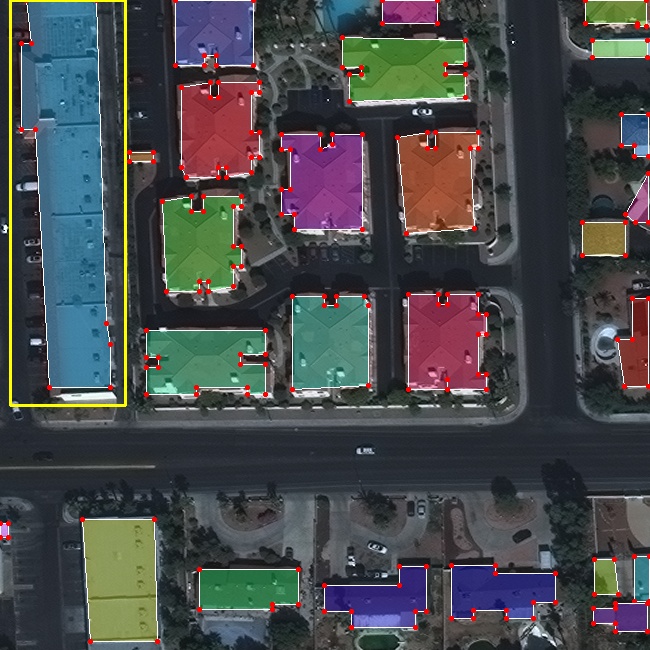}
		\end{minipage}
		\begin{minipage}[t]{0.23\linewidth}
			\centering
			\includegraphics[width=1\linewidth]{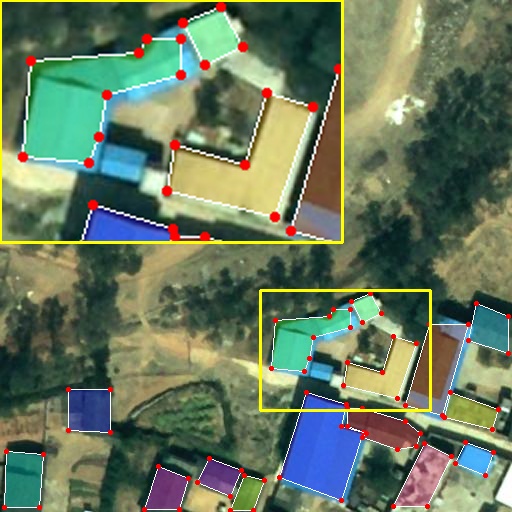}
		\end{minipage}
		\begin{minipage}[t]{0.23\linewidth}
			\centering
			\includegraphics[width=1\linewidth]{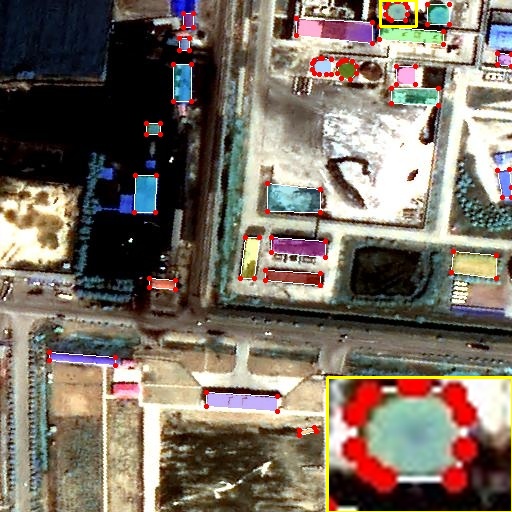}
		\end{minipage}
	}
	\caption{Qualitative results. Our method can generate geometric contours of buildings accurately.}
	\label{Fig3}
\end{figure}

\noindent \textbf{Qualitative Comparison}.
Figure \ref{Fig3} shows some example results obtained by our approach. It can be seen that our method can generate high-quality polygonal building footprints.

\begin{table*} [htbp]
\renewcommand\arraystretch{0.85}
\centering
\scalebox{0.9}{
\begin{tabular}{c|c|ccc|ccc|c}
\hline
Dataset &Method      & AP & $AP_{50}$ & $AP_{75}$ & AR & $AR_{50}$ & $AR_{75}$ & $F1_{75}$ \\
\hline
\multirow{4}*{\makecell[c]{SpaceNet \\ (LasVegas)}} & PANet \cite{liu2018panet} & 46.9  & 85.1  & 45.9  & 54.6  & 87.8  & 60.9 & 52.35     \\
~ & PolyMapper \cite{li2019polymapper} & 51.6 & \textbf{87.4} & 59.6 & 58.3 & \textbf{89.5} & 68.9 & 63.91 \\
~ & FrameField \cite{girard2021framefield} & \textbf{53.6} & 84.5 & \textbf{63.1} & 58.5 & 87.9 & 68.8 & 65.83  \\
~ & Baseline \cite{he2017maskrcnn} & 47.0  & 85.9  & 46.4 & 54.6  & 88.0  & 60.5 & 52.52   \\
~ & BiSVP (ours) & $53.2_{+6.2}$ & $87.2_{+1.3}$ & $62.1_{+15.7}$ & $\textbf{59.3}_{+4.7}$ & $\textbf{89.5}_{+1.5}$ & $\textbf{70.1}_{+9.6}$ & $\textbf{65.86}_{+13.34}$ \\
\hline
\multirow{4}*{5M-Building} & PANet \cite{liu2018panet} & 31.3  & 59.6  & 28.9  & 45.8  & 74.7  & 48.5 & 36.22     \\
~ & PolyMapper \cite{li2019polymapper} & 32.0 & 62.9 & 30.1 & 47.5 & 82.1 & 51.1 & 37.88 \\
~ & FrameField \cite{girard2021framefield} & 18.4 & 36.4 & 16.4 & 31.0 & 54.7 & 31.0 & 21.45  \\
~ & Baseline \cite{he2017maskrcnn} & 31.5  & 60.5  & 29.0  & 45.9  & 75.4  & 48.3 & 36.24   \\
~ & BiSVP (ours) & $\textbf{32.7}_{+1.2}$ & $\textbf{63.8}_{+3.3}$ & $\textbf{31.1}_{+2.1}$ & $\textbf{48.7}_{+2.8}$ & $\textbf{83.9}_{+8.5}$ & $\textbf{52.9}_{+4.6}$ & $\textbf{39.17}_{+2.93}$ \\
\hline
\multirow{4}*{CNData} & PANet \cite{liu2018panet} & 35.1  & 68.8  & 34.0  & 47.5  & 81.3  & 50.3 & 40.57     \\
~ & PolyMapper \cite{li2019polymapper} & 36.4 & 70.6 & 35.7 & 50.6 & 86.1 & 54.6 & 43.17  \\
~ & FrameField \cite{girard2021framefield} & 21.7 & 40.7 & 21.2 & 32.9 & 54.9 & 34.4 & 26.23  \\
~ & Baseline \cite{he2017maskrcnn} & 35.1  & 68.4  & 33.7  & 47.7  & 81.6  & 50.2 & 40.33   \\
~ & BiSVP (ours) & $\textbf{37.9}_{+2.8}$ & $\textbf{71.5}_{+3.1}$ & $\textbf{37.7}_{+4.0}$ & $\textbf{52.7}_{+5.0}$ & $\textbf{88.3}_{+6.7}$ & $\textbf{57.3}_{+7.1}$ & $\textbf{45.48}_{+5.15}$  \\
\hline
\end{tabular}
}
\caption{Reuslts on three building test datasets: SpaceNet (LasVegas), 5M-Building, and CNData. The best results in each dataset group are marked in bold.}
\label{tab1}
\end{table*}
\vspace{-10pt}
\subsection{Ablation Study}
We analyze the effectiveness of CSFF module, Bi-BP module, and the attention mechanism (${\text{Attn}}_{gc}$) in Bi-BP. In ablation studies, we add CSFF, Bi-BP, and ${\text{Attn}}_{gc}$ respectively to the baseline method \cite{he2017maskrcnn}. The experimental results are present in Table \ref{tab2}.

\noindent \textbf{Cross-scale feature fusion (CSFF)}.
The performance of the baseline model decreases on the three building datasets by 12.27\%, 2.26\%, and 4.73\% in the indicator $F1_{75}$. The results shown in Table \ref{tab2} indicate that CSFF module plays a vital role in aggregating features of different levels.

\noindent \textbf{Bidirectional building polygon (Bi-BP)}.
Table \ref{tab2} shows that Bi-BP module can significantly improve the performance on the three building datasets, which demonstrates that Bi-BP module can leverage the \textit{bidirectional} information of the building polygon. Especially, it improves the baseline by 12.8\% in $F1_{75}$ of SpaceNet (Las Vega), proving the effectiveness of the Bi-BP module.

\noindent \textbf{Attention mechanism}.
We can observe from Table \ref{tab2} that the model with the attention mechanism can achieve surprisingly good performance. Finally, the proposed BiSVP can improve the baseline from 52.52\%, 36.24\%, and 40.33\% to 65.86\% (+13.34\%), 39.17\% (+2.93\%), and 45.48\% (+5.15\%) in $F1_{75}$ over three building test datasets, respectively.

\begin{table} [!tb]
\renewcommand\arraystretch{0.8}
\centering
\begin{tabular}{c|ccc|c}
\hline
Dataset & CSFF & Bi-BP & Attn & $F1_{75}$ \\
\hline
\multirow{5}*{\makecell[c]{SpaceNet \\ (LasVegas)}} &  &  &  & 52.52   \\
~ & \checkmark &  &  & 64.79     \\
~ &  & \checkmark &  & 65.32 \\
~ &  &  & \checkmark & 65.44 \\
~ & \checkmark & \checkmark & \checkmark & \textbf{65.86} \\
\hline
\multirow{5}*{5M-Building} &  &  &  & 36.24   \\
~ & \checkmark &  &  & 38.50     \\
~ &  & \checkmark &  & 38.58 \\
~ &  &  & \checkmark & 39.03 \\
~ & \checkmark & \checkmark & \checkmark & \textbf{39.17} \\
\hline
\multirow{5}*{CNData} &  &  &  & 40.33   \\
~ & \checkmark &  &  & 45.06     \\
~ &  & \checkmark &  & 43.35 \\
~ &  &  & \checkmark & 44.80 \\
~ & \checkmark & \checkmark & \checkmark & \textbf{45.48}  \\
\hline
\end{tabular}

\caption{Ablation study. "\checkmark" means adding the corresponding module to the baseline. The last row in each dataset group is the value of BiSVP. The best result in each dataset group is marked in bold.}
\label{tab2}
\end{table}

\section{Conclusion}
\label{sec:conclusion}

In this paper, we have presented Bidirectional Serialized Vertex Prediction (BiSVP), a new refinement-free and end-to-end framework to extract building footprints from remote sensing images. The proposed BiSVP represents the building contour with ordered vertices and predicts the serialized vertices directly in a bidirectional fashion. Furthermore, BiSVP proposes a cross-scale feature fusion (CSFF) module to fuse feature maps of different levels, obtaining the building feature with rich spatial and context information that is essential for the dense building vertex prediction task. The extensive experiments on three building instance segmentation datasets demonstrate the superiority of our method in building footprint extraction.

\vfill\pagebreak

{\small
\bibliographystyle{IEEEbib}
\bibliography{reference}
}
\end{document}